\documentclass{article}

\usepackage[preprint]{neurips_2025}

\usepackage[utf8]{inputenc} 
\usepackage[T1]{fontenc}    
\usepackage{hyperref}       
\usepackage{url}            
\usepackage{booktabs}       
\usepackage{amsfonts}       
\usepackage{nicefrac}       
\usepackage{microtype}      
\usepackage{xcolor}         
\usepackage{graphicx}
\usepackage{amsmath}
\usepackage{graphicx}
\usepackage{subfigure}

\title{DeCoDe: Defer-and-Complement Decision-Making via Decoupled Concept Bottleneck Models}

\author{%
  Chengbo He \\
  University of Science and Technology Beijing \\
  Beijing, China \\
  \And
  Bochao Zou \\
  University of Science and Technology Beijing \\
  Beijing, China \\
  \And
  Junliang Xing \\
  Tsinghua University \\
  Beijing, China \\
  \And
  Jiansheng Chen \\
  University of Science and Technology Beijing \\
  Beijing, China \\
  \And
  Yuanchun Shi \\
  Tsinghua University \\
  Beijing, China \\
  \And
  Huimin Ma \\
  University of Science and Technology Beijing \\
  Beijing, China \\
}

\begin{document}

\maketitle

\begin{abstract}

In human-AI collaboration, a central challenge is deciding whether the AI should handle a task, be deferred to a human expert, or be addressed through collaborative effort. Existing Learning to Defer approaches typically make binary choices between AI and humans, neglecting their complementary strengths. They also lack interpretability, a critical property in high-stakes scenarios where users must understand and, if necessary, correct the model’s reasoning. To overcome these limitations, we propose \textbf{De}fer-and-\textbf{Co}mplement Decision-Making via \textbf{De}coupled Concept Bottleneck Models \textbf{(DeCoDe)}, a concept-driven framework for human-AI collaboration. DeCoDe makes strategy decisions based on human-interpretable concept representations, enhancing transparency throughout the decision process. It supports three flexible modes: autonomous AI prediction, deferral to humans, and human-AI collaborative complementarity, selected via a gating network that takes concept-level inputs and is trained using a novel surrogate loss that balances accuracy and human effort. This approach enables instance-specific, interpretable, and adaptive human-AI collaboration. Experiments on real-world datasets demonstrate that DeCoDe significantly outperforms AI-only, human-only, and traditional deferral baselines, while maintaining strong robustness and interpretability even under noisy expert annotations. 

\end{abstract}

\section{Introduction}
\label{Introduction}

Human-AI collaboration is increasingly adopted to tackle high-stakes, real-world decision-making tasks where neither party alone is fully reliable \citep{mozannar2023should, gao2024taxonomy, hemmer2024complementarity}. To enhance robustness and safety, machine learning models are often endowed with the ability to \textit{defer decisions} to humans when faced with unfamiliar or uncertain cases \citep{mozannar2023effective}. This strategy, known as \textit{Learning to Defer} (L2D) \citep{madras2018predict}, enables the model to learn when to predict jointly and when to defer. Compared to AI-only or human-only approaches, such collaboration yields better trade-offs between accuracy and trustworthiness \citep{marusich2024using, wei2024exploiting}, especially in sensitive domains such as medical imaging \citep{chang2025artificial, halling2020optimam, frazer2024comparison}, clinical risk assessment \citep{dvijotham2023enhancing, mao2025multimodal}, and harmful content detection in LLMs \citep{rajashekar2024human, he2024enhancing}.

\begin{figure}[t]
    \centering
    \includegraphics[width=0.9\linewidth]{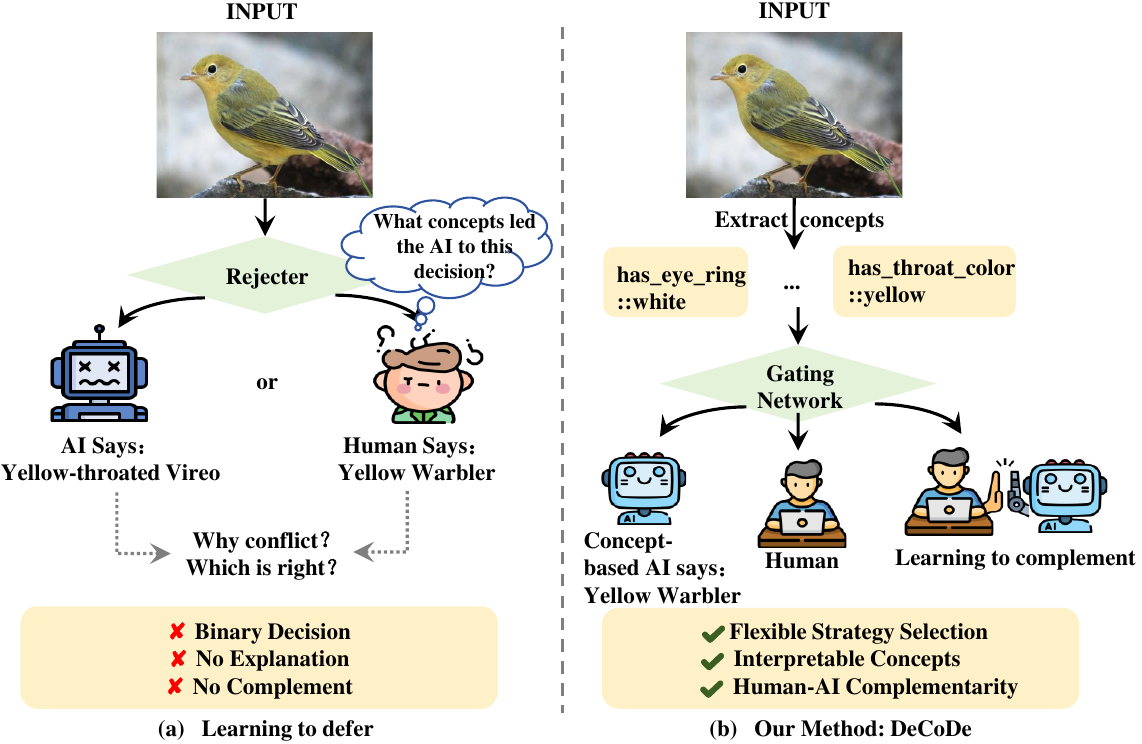}
    \caption{
    \textbf{Comparison between traditional Learning to Defer (left) and our proposed DeCoDe framework (right)}. 
    While conventional deferral methods make binary AI–human decisions without providing explanations or enabling collaboration, DeCoDe leverages interpretable concept representations to support flexible strategy selection—autonomous AI, human-only, and human-AI complementarity—guided by a concept-driven gating network.
    }
    \label{figure1}
\end{figure}

While L2D has shown strong empirical performance across tasks by allowing AI systems to seek human input under uncertainty, current frameworks still face critical limitations in strategy expressiveness, semantic interpretability, and robustness to human variability \citep{nguyen2025probabilistic, zhang2024learning, bansal2021does}. 

First, most L2D approaches restrict the model to binary strategies—either fully autonomous prediction or complete deferral to humans—limiting the ability to effectively leverage complementary strengths of humans and AI within the same instance, which may hinder the potential for hybrid decision-making strategies that combine both \citep{donahue2024two, narasimhan2022post, verma2023learning, jarrahi2022artificial}. 

Second, deferral decisions typically rely on latent internal signals such as confidence or hidden representations \citep{madras2018predict, wu2025learning, mozannar2020consistent}. While useful for optimization, these signals lack semantic transparency, making it difficult for users to understand why the model defers—a crucial issue in high-stakes settings \citep{jiang2022needs}. 

Finally, many frameworks implicitly assume humans are inherently more reliable, treating deferral as a safe fallback. However, human performance varies in practice due to differences in expertise, cognitive styles, and task familiarity \citep{glickman2025human, pataranutaporn2023influencing}. Rigid, one-size-fits-all deferral strategies can therefore degrade performance and erode user trust.

To address the above challenges, we propose \textbf{De}fer-and-\textbf{Co}mplement Decision-Making via \textbf{De}coupled Concept Bottleneck Models \textbf{(DeCoDe)}. This human-AI collaboration framework combines semantic interpretability with flexible strategy selection (see Figure~\ref{figure1}). DeCoDe adopts a concept-driven modeling approach that uses explicitly labeled human concepts as intermediate representations \citep{zhang2024decoupling}. These semantically structured concepts provide a shared foundation for classification and strategy selection, enhancing transparency and avoiding reliance on latent features. The concept layer improves interpretability and enables controllable, concept-based collaboration. DeCoDe dynamically allocates decision weights between AI and humans based on available concept information \citep{zhang2024learning, zhang2024coverage}, supporting three collaboration modes: autonomous AI prediction, deferral to humans, and human-AI complementarity. This design avoids the common issue of blindly deferring to humans in existing L2D methods. When human labels are noisy or unreliable, DeCoDe evaluates the suitability of each strategy to prevent performance degradation. By dynamically adjusting strategy choices on a per-instance basis, DeCoDe improves decision quality, robustness, and adaptability in complex real-world settings.

This paper makes the following three main contributions:
\begin{itemize}
\item We propose DeCoDe, a human-AI collaboration framework that uses explicit concept representations to support classification and adaptive strategy selection.
\item We design a concept-based gating network trained with a novel surrogate loss, enabling per-instance selection among AI-only, human-only, and AI-human collaboration.
\item DeCoDe assigns strategy weights over concept space to prevent performance degradation from low-quality human input, improving system robustness and adaptability.
\end{itemize}
With these contributions, we have obtained a system that performs semantically interpretable and instance-adaptive collaboration between AI and human decision-makers. By dynamically selecting among three decision strategies using concept-based reasoning, the DeCoDe system not only reduces unnecessary human workload but also avoids performance degradation caused by unreliable human input, a standard failure mode in L2D methods. Extensive experiments on real-world datasets, such as CUB-200-2011~\citep{welinder2010caltech}, Derm7pt~\citep{kawahara2018seven}, and CelebA~\citep{liu2015deep}, show that DeCoDe achieves superior trade-offs between accuracy, human cost, and robustness, even under noisy or inconsistent human supervision.

\section{Related work}
\label{Related work}

\textbf{Learning to Defer for Human-AI Collaboration}\quad
Learning to Defer (L2D) frameworks aim to optimize task allocation between AI and human experts by allowing the system to defer uncertain cases to the human \citep{madras2018predict, mozannar2020consistent}. Early works estimate human error or confidence to determine whether to defer \citep{raghu2019algorithmic, madras2018predict}, while later methods formulate deferral as a differentiable objective and introduce surrogate losses to support end-to-end learning \citep{mozannar2020consistent, wilder2021learning, narasimhan2022post}. However, most L2D frameworks rely on latent features or black-box confidence signals, offering little interpretability and limited controllability. In addition, they typically assume humans are more reliable and restrict decision-making to a binary choice between full AI prediction and full deferral. \citet{zhang2024learning} attempts to unify deferral and complementarity, but treats collaboration as a label denoising problem and still operates on latent features. These limitations make it difficult to reason when and why the model differs. In contrast, DeCoDe performs strategy selection over explicit semantic concepts, enabling transparent and adaptive collaboration across three modes: AI-only, human-only, and AI-human complementarity.

\vspace{1em}

\textbf{Concept Bottleneck Models for Interpretable Decision-Making}\quad
Concept Bottleneck Models (CBMs) enhance model transparency by introducing human-understandable concepts as intermediate representations \citep{koh2020concept}. CBMs predict concepts first and use them to infer the final label, allowing semantic-level inspection and intervention. However, they often assume a complete and sufficient concept set for accurate prediction, which may fail in real-world scenarios with latent or unmodeled factors.
\citep{kazhdan2021disentanglement, yuksekgonul2022post}. When this assumption breaks, task-relevant information may leak into the concept space, reducing interpretability. Several extensions have been proposed to address these issues, such as the decoupling concept and task information \citep{zhang2024decoupling}, introducing auxiliary channels \citep{havasi2022addressing}, and enabling tree-based concept editing \citep{ragkousis2024tree}. However, these methods still lack a principled mechanism for handling instances that remain uncertain despite accurate concept predictions, and they do not account for how humans can complement AI when concept-level reasoning alone is insufficient. DeCoDe builds on decoupled CBMs by introducing a concept-aware strategy selector that dynamically chooses whether to defer, collaborate, or predict autonomously, preserving interpretability while improving adaptability in uncertain, high-stakes settings.

\begin{figure}[t]
    \centering
    \includegraphics[width=1\linewidth]{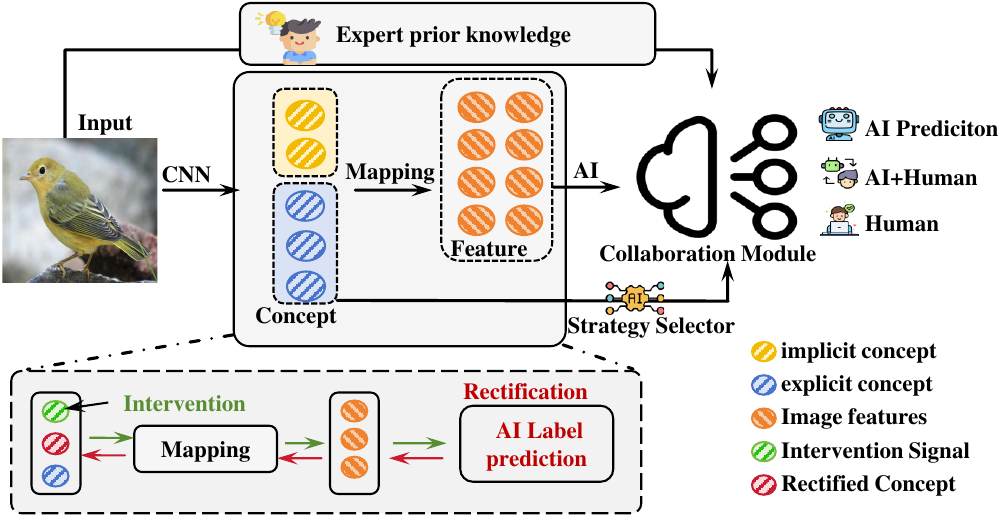}
    \caption{
    \textbf{Overall architecture of the DeCoDe framework.} The model is built upon a concept bottleneck structure, where explicit concepts are extracted from input images to form intermediate representations. These representations are used for downstream task prediction and strategy selection, enabling the model to adaptively choose among three decision modes: autonomous AI prediction, deferral to humans, and AI-human collaboration. The dashed box at the bottom illustrates the interactive mechanism of the system: users can intervene on the concept layer to correct the prediction or adjust the concept representation by rectifying the model’s output, thereby enhancing interpretability and controllability.
    }
    \label{figure2}
\end{figure}

\section{Method}
\label{Method}

This section details the proposed \textbf{DeCoDe} framework, with the overall system pipeline illustrated in Figure~\ref{figure2}. The input image is first processed by the \textit{CNN} to extract an explicit concept vector $c_{\text{exp}}$ and an implicit latent representation $c_{\text{imp}}$ (Section~\ref{sec:architecture}). The \textit{Strategy Selector} then takes $c_{\text{exp}}$ as input and outputs a probability distribution over three collaboration modes: AI-only, AI+Human, and Defer-to-Human (Section~\ref{sec:strategy}). Based on this distribution, the \textit{Collaboration Module} integrates predictions from the model and the expert to produce the final output. Section~\ref{sec:Loss} introduces the surrogate loss $\mathcal{L}_{\text{DeCoDe}}$, which enables joint optimization of the strategy selector and collaboration mechanism in the absence of explicit routing labels.

\subsection{Concept-Based Model Architecture}
\label{sec:architecture}

DeCoDe is a concept-driven framework built upon the CBM paradigm \citep{koh2020concept}, which effectively leverages intermediate concept-level representations to significantly enhance both interpretability and controllability. The training data consists of $N$ triplets $\{(x_n, c_n, y_n)\}_{n=1}^N$, where $x_n \in \mathbb{R}^D$ is the input feature, $c_n \in \mathbb{R}^d$ is a $d$-dimensional vector of human-annotated concepts, and $y_n \in \mathbb{R}^k$ is the corresponding task label.

A standard CBM follows a two-stage structure: a concept predictor $g: \mathbb{R}^D \rightarrow \mathbb{R}^d$ first maps the input features to the concept space, and a classifier $f: \mathbb{R}^d \rightarrow \mathbb{R}^k$ then produces the final prediction based on these concept-level representations. The overall prediction process is:

\begin{equation}
P(y \mid x) = f(g(x)) = f(c).
\label{eq:cbm}
\end{equation}

This structure confines the decision process entirely to the interpretable concept space. However, in real-world settings, concept annotations are often incomplete and insufficient to capture all predictive factors. To balance interpretability and representational capacity, DeCoDe builds on the decoupled structure proposed in DCBM \citep{zhang2024decoupling}, which separates the intermediate representation into an explicit concept vector $c_{\text{exp}}$ and an implicit latent vector $c_{\text{imp}}$. The final prediction jointly incorporates both components:
\begin{equation}
P(y \mid x) = f(c_{\text{exp}}) + \tilde{f}(c_{\text{imp}}),
\label{eq:decompose}
\end{equation}
where \( f(c_{\text{exp}}) \) models the interpretable component of the output, while \( \tilde{f}(c_{\text{imp}}) \) supplements it with task-relevant latent features beyond the concept space.

To encourage balanced reliance on both components, a distributional consistency constraint is applied to prevent the model from entirely ignoring the explicit concept signal. Specifically, the Jensen–Shannon divergence is used to align the interpretable prediction with the full output. The corresponding loss function is:

\begin{equation}
\mathcal{L}_{\text{JS}} = \text{JS}\left(f(g(x)) \ \| \ f(g(x)) + \tilde{f}(\tilde{g}(x))\right).
\label{eq:js_loss}
\end{equation}

The concept representation module is then treated as a fixed feature extractor for downstream components, providing both interpretable concept vectors and predictive logits to support strategy selection and human-AI collaboration.

\subsection{Concept-Guided Strategy Allocation and Human-AI Fusion}
\label{sec:strategy}

Based on the explicit concept representations introduced earlier, DeCoDe incorporates a strategy selection module and a collaboration module to dynamically determine the decision path for each instance—whether the prediction should be made solely by the model, jointly by the model and a human, or entirely deferred to the expert. We refer to the three collaborative modes as \textit{AI-only}, \textit{AI+Human}, and \textit{Defer-to-Human}. The system selects between these modes to balance predictive performance and human resource usage.

The strategy selection module is implemented as a three-way classifier that takes the concept representation $c_{\text{exp}}$ as input and outputs a strategy distribution. Formally, we define $g_\phi: \mathcal{C} \rightarrow \Delta^3$, where $\mathcal{C}$ denotes the space of explicit concept representations and $c_{\text{exp}} \in \mathcal{C}$. The target space $\Delta^3$ is the 3-dimensional probability simplex representing the distribution over three strategies.

The collaboration module $h_\psi$ receives the strategy distribution $g_\phi(c_{\text{exp}})$, the model prediction $f_\theta(x)$, and the expert label $m_i$, and dynamically constructs the final input based on the most probable strategy. The function $\mathsf{p}(\cdot)$ selects the appropriate combination of model and expert predictions as follows:
\begin{equation}
    \mathsf{p}(g_\phi(c_{\text{exp}}), f_\theta(x), m_i) =
    \begin{cases}
        [f_\theta(x), \mathbf{0}] & \text{if } \arg\max_j g_\phi^{(j)}(c_{\text{exp}}) = 1 , \\
        [f_\theta(x), m_i] & \text{if } \arg\max_j g_\phi^{(j)}(c_{\text{exp}}) = 2 , \\
        [\mathbf{0}, m_i] & \text{if } \arg\max_j g_\phi^{(j)}(c_{\text{exp}}) = 3 ,
    \end{cases}
\end{equation}
where $\mathbf{0}$ is a zero vector with the same dimension as the model output. This design enables adaptive integration of expert input during prediction, enhancing flexibility and transparency.

DeCoDe employs a unified model architecture and introduces a surrogate loss $\mathcal{L}_{\text{co}}$ to optimize strategy selection and collaborative prediction jointly. In contrast to LECODU \citep{zhang2024learning}, which trains two separate AI models with cross-generated pseudo-labels, DeCoDe avoids inter-model consistency assumptions and reduces structural complexity. Strategy supervision is based on per-sample comparisons between model and expert predictions, from which soft pseudo-labels are constructed to enhance robustness and adaptability.

The overall training objective balances prediction accuracy with the cost of using human expertise:
\begin{equation}
    \mathcal{L}_{\text{total}} = \mathcal{L}_{\text{co}} + \lambda \cdot \text{cost}(g_\phi(x)).
\end{equation}
Here, $\mathcal{L}_{\text{co}}$ supervises both the strategy selector $g_\phi$ and the collaboration module $h_\psi$, and $\lambda$ is a tunable hyperparameter that penalizes human involvement.

The cost term is defined using $g_\phi^{(j)}(x)$, the predicted probability of selecting the $j$-th strategy, where $j \in \{1, 2, 3\}$ corresponds to \textit{AI-only}, \textit{AI+Human}, and \textit{Defer-to-Human}, respectively. In the single-expert setting, DeCoDe applies a simple additive penalty for any strategy that involves the expert:
\begin{equation}
    \text{cost}(g_\phi(x)) = g_\phi^{(2)}(x) + g_\phi^{(3)}(x).
    \label{eq:cost}
\end{equation}
This formulation imposes no penalty when the model acts independently ($j = 1$), and applies a unit cost when the strategy includes human input through either collaboration or deferral.

Our objective offers three key advantages: interpretability, through concept-based strategy selection; controllability, via cost-regularized expert usage; and flexibility, by modeling strategies in a continuous probabilistic space for adaptive decision-making. The following section describes the formulation and optimization of $\mathcal{L}_{\text{co}}$ in detail.

\subsection{Surrogate Loss Design and Optimization}
\label{sec:Loss}

Optimizing the strategy selector $g_\phi$ is challenging due to the absence of ground-truth routing labels. To address this, we propose a surrogate loss that encourages alignment between the predicted strategy distribution $r(x)$ and a rule-based pseudo-label $q(x)$ derived from the relative correctness of the AI and human experts. The strategy selector takes concept-level representations $c_{\text{exp}} = g(x)$ as input, providing a semantically interpretable basis for decision supervision.

For each instance, the target strategy is determined using a heuristic rule: the model selects \textit{AI-only} when the AI prediction is correct, \textit{Defer-to-Human} if at least one expert is correct while the AI is not, and defaults to \textit{AI+Human} otherwise. This rule produces a pseudo-label $\tilde{s}(x) \in \mathcal{S}$, which can be converted to a hard supervision vector:
\begin{equation}
q^{(s)}(x) = \mathbb{I}[s = \tilde{s}(x)], \quad \forall s \in \mathcal{S}.
\label{eq:hard_q}
\end{equation}

Optionally, we construct a soft variant to reflect uncertainty in strategy suitability. This is computed by applying a softmax over the negative heuristic losses:
\begin{equation}
q^{(s)}(x) = \frac{\exp(-\ell^{(s)}(x))}{\sum_{s' \in \mathcal{S}} \exp(-\ell^{(s')}(x))},
\label{eq:soft_q}
\end{equation}
where $\ell^{(s)}(x)$ is a behavior-driven penalty for selecting strategy $s$ (see Appendix A).

The overall surrogate loss balances two objectives: learning an appropriate strategy selection and ensuring a correct final prediction. Formally, we write:
\begin{equation}
\mathcal{L}_{\text{DeCoDe}}(x, y) = \underbrace{\mathcal{L}_{\text{strategy}}(r(x), q(x))}_{\text{strategy supervision}} + \alpha \cdot \underbrace{\mathcal{L}_{\text{CE}}(\hat{y}(x), y)}_{\text{classification loss}},
\label{eq:surrogate_total}
\end{equation}
where $\hat{y}(x)$ is the final output of the collaboration module $h_\psi$, and $\alpha$ controls the relative importance of the two components in the overall decision-making process. Our surrogate loss supports both hard and soft supervision, instantiated as cross-entropy or KL divergence depending on the form of the pseudo-label $q(x)$ (see Appendix B for further details and derivations).

\section{Experimental}
\label{Experimental}

\subsection{Experimental Setup}
\label{sec:experimental_setup}

\textbf{Datasets.}  
We use CUB-200-2011~\citep{welinder2010caltech}, Derm7pt~\citep{kawahara2018seven}, and CelebA~\citep{liu2015deep}, which provide structured semantic concepts suitable for interpretable modeling. As these datasets do not include human decisions, we simulate expert annotations by injecting label noise at rates ranging from 10\% to 50\%, representing different levels of human reliability. This simulation follows previous work~\citep{zhang2024learning} and is commonly used in the field to model human decision-making under noisy conditions. Dataset statistics are summarized in Table~\ref{tab:dataset_summary}, with additional details provided in Appendix C.

\textbf{Architecture.}  
All models are implemented in PyTorch and trained on NVIDIA RTX A6000 GPUs. The concept encoder $g(x)$ maps input images to semantic vectors $c$ and shares the same backbone as the classifier $f(c)$ to ensure representational consistency. We use Inception-v3 \citep{szegedy2016rethinking} for CUB and Derm7pt, and ResNet-18 \citep{he2016deep} for CelebA. The concept-to-label mapping is implemented as a single-layer linear classifier, preserving interpretability and allowing semantic-level interventions for improved model transparency.

\textbf{Model Variants and Baselines.}  
We consider two versions of DeCoDe: \textit{DeCoDe-Defer-human}, which supports only the “AI-only” and “Defer-to-Human” strategies; and \textit{DeCoDe}, which additionally models “AI+Human” collaboration. We compare against three representative L2D baselines: Cross-Entropy surrogate (CE)~\citep{mozannar2020consistent}, One-vs-All surrogate (OvA)~\citep{verma2022calibrated}, and Realizable Surrogate (RS)~\citep{mozannar2023should}. In addition, we include stand-alone AI and simulated human-only predictions to establish upper and lower performance bounds. All methods are built upon a shared DCBM to ensure architectural fairness.

\textbf{Evaluation and Metrics.}  
Expert labels are simulated by injecting instance-level label noise at predefined rates. Collaboration cost is measured by the number of expert queries incurred at test time. We report top-1 classification and concept prediction accuracy and visualize the trade-off between performance and expert usage using accuracy–coverage curves. To ensure a fair and consistent comparison, all models share the same DCBM. Our objective is not to improve raw accuracy through architectural enhancements but to evaluate interpretable and controllable human–AI collaboration grounded in concept-level reasoning.

\begin{table}[t]
\centering
\caption{Brief Summary of the Datasets}
\label{tab:dataset_summary}
\small
\resizebox{0.95\textwidth}{!}{%
\begin{tabular}{lcccccc}
\toprule
\textbf{Dataset} & \textbf{Concept} & \textbf{Class} & \textbf{Train} & \textbf{Val} & \textbf{Test} \\
\midrule
CUB~\cite{welinder2010caltech}      & 112 & 200 & 4796 & 1198 & 5794 \\

Derm7pt~\cite{kawahara2018seven}    & 8   & 2    & 688  & 322  & 636  \\

CelebA~\cite{liu2015deep}           & 6   & 256  & 11818 & 1689 & 3376 \\
\bottomrule
\end{tabular}%
}
\end{table}

\subsection{Results}
\label{Results}

\begin{figure}[t]
  \centering
  \subfigure[CUB, Noise 0.1 \label{fig:cub_noise_01}]{
    \includegraphics[width=0.3\linewidth]{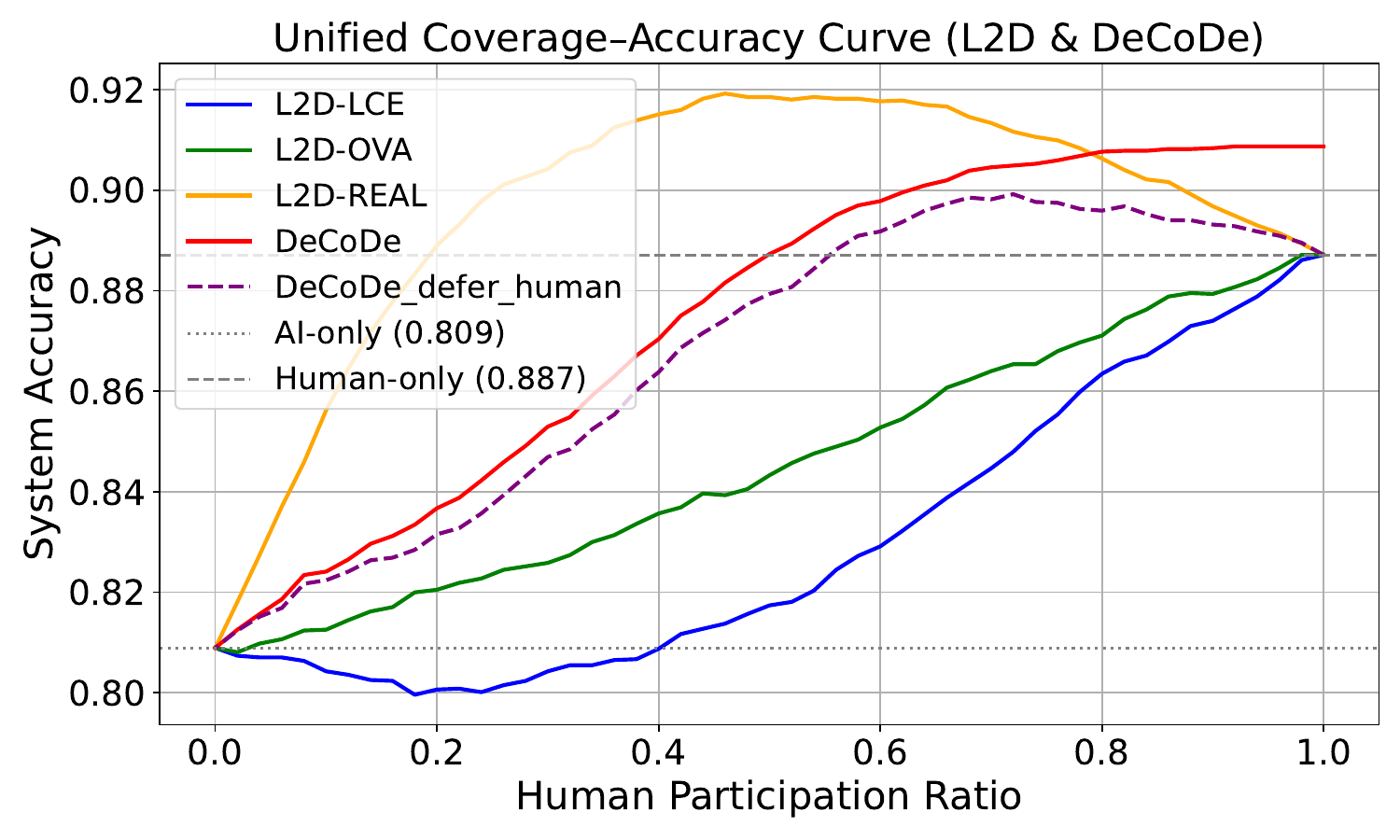}
  }
  \hspace{0.5em}
  \subfigure[CUB, Noise 0.3 \label{fig:cub_noise_03}]{
    \includegraphics[width=0.3\linewidth]{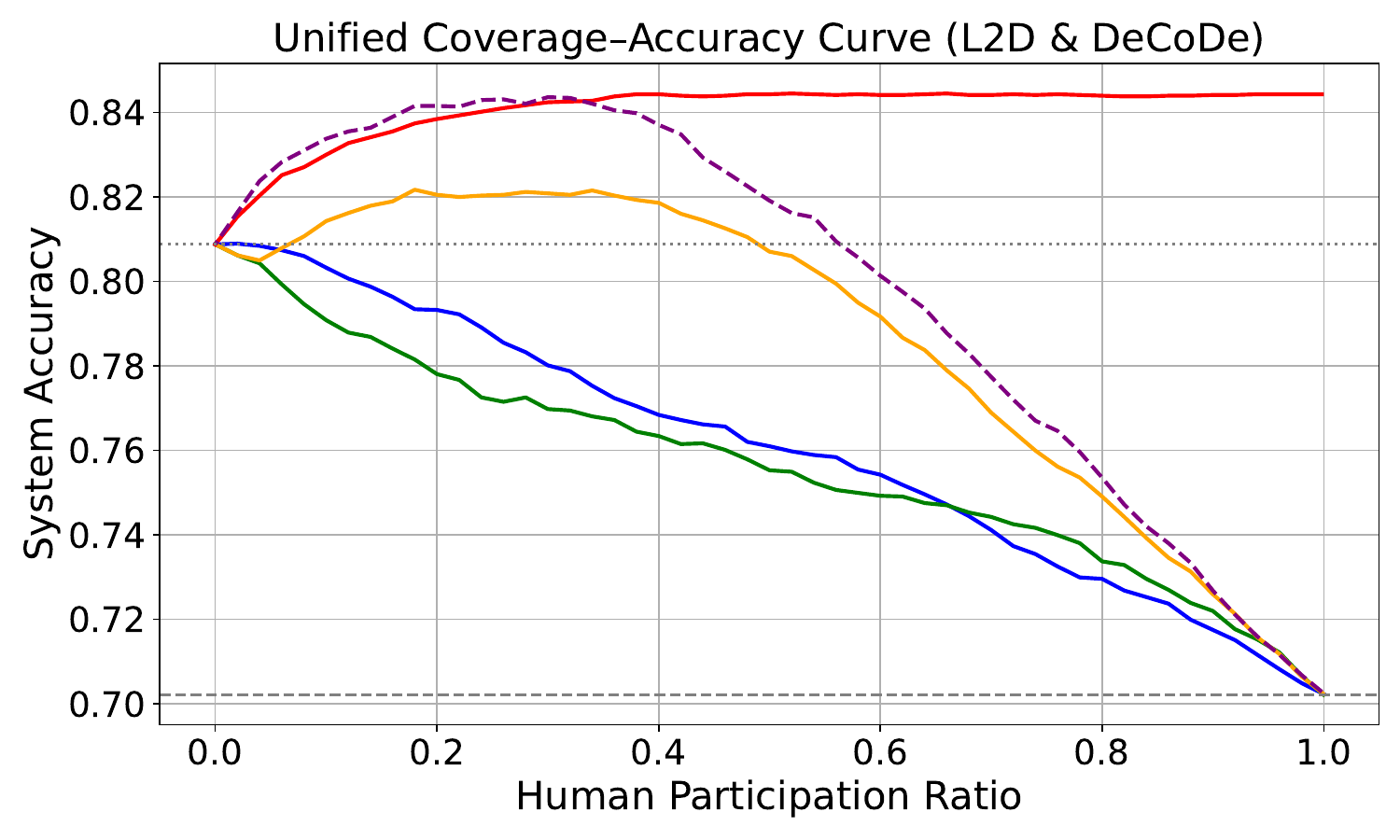}
  }
  \hspace{0.5em}
  \subfigure[CUB, Noise 0.5 \label{fig:cub_noise_05}]{
    \includegraphics[width=0.3\linewidth]{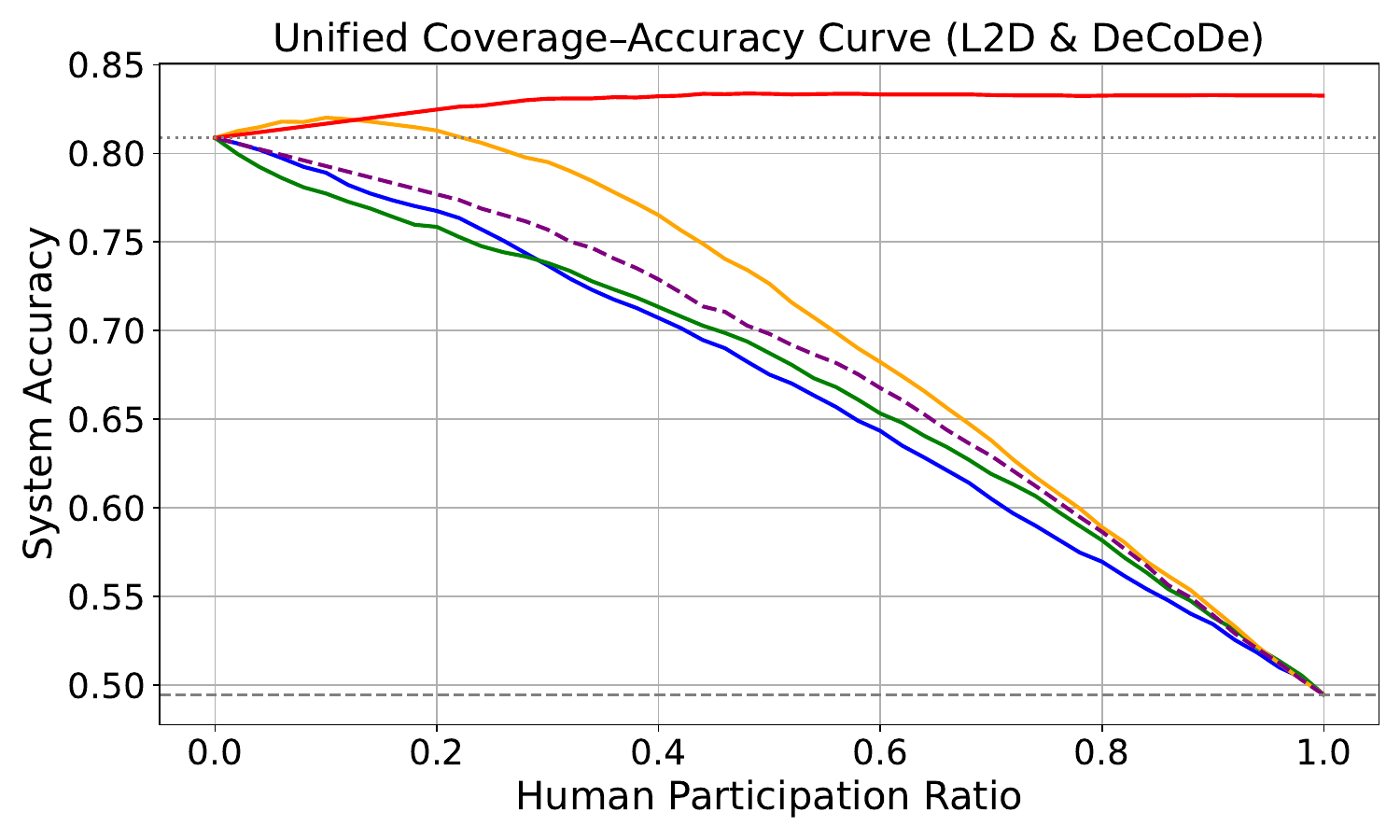}
  }
  \vspace{-0.5em}
  \subfigure[Derm7pt, Noise 0.1 \label{fig:derm_noise_01}]{
    \includegraphics[width=0.3\linewidth]{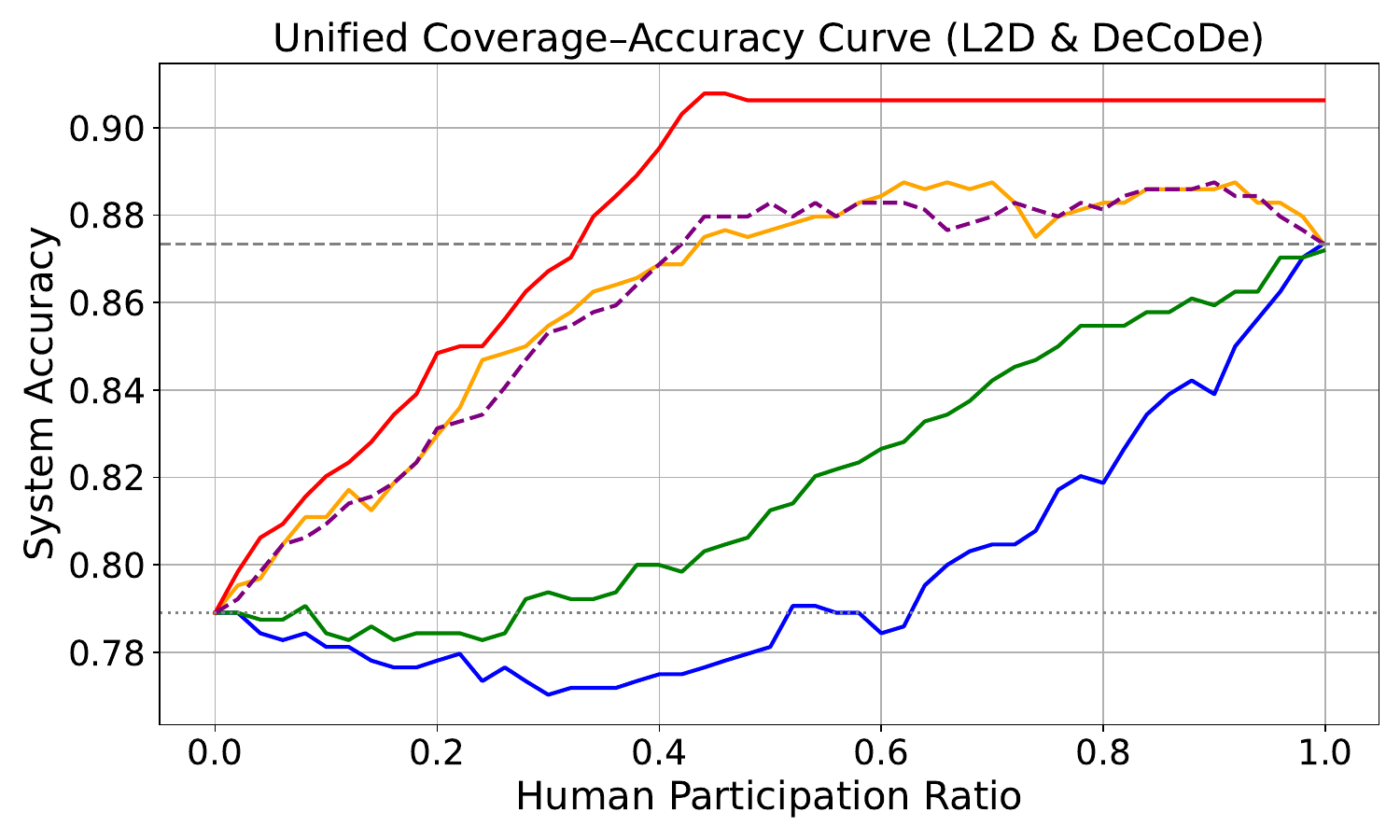}
  }
  \hspace{0.5em}
  \subfigure[Derm7pt, Noise 0.3 \label{fig:derm_noise_03}]{
    \includegraphics[width=0.3\linewidth]{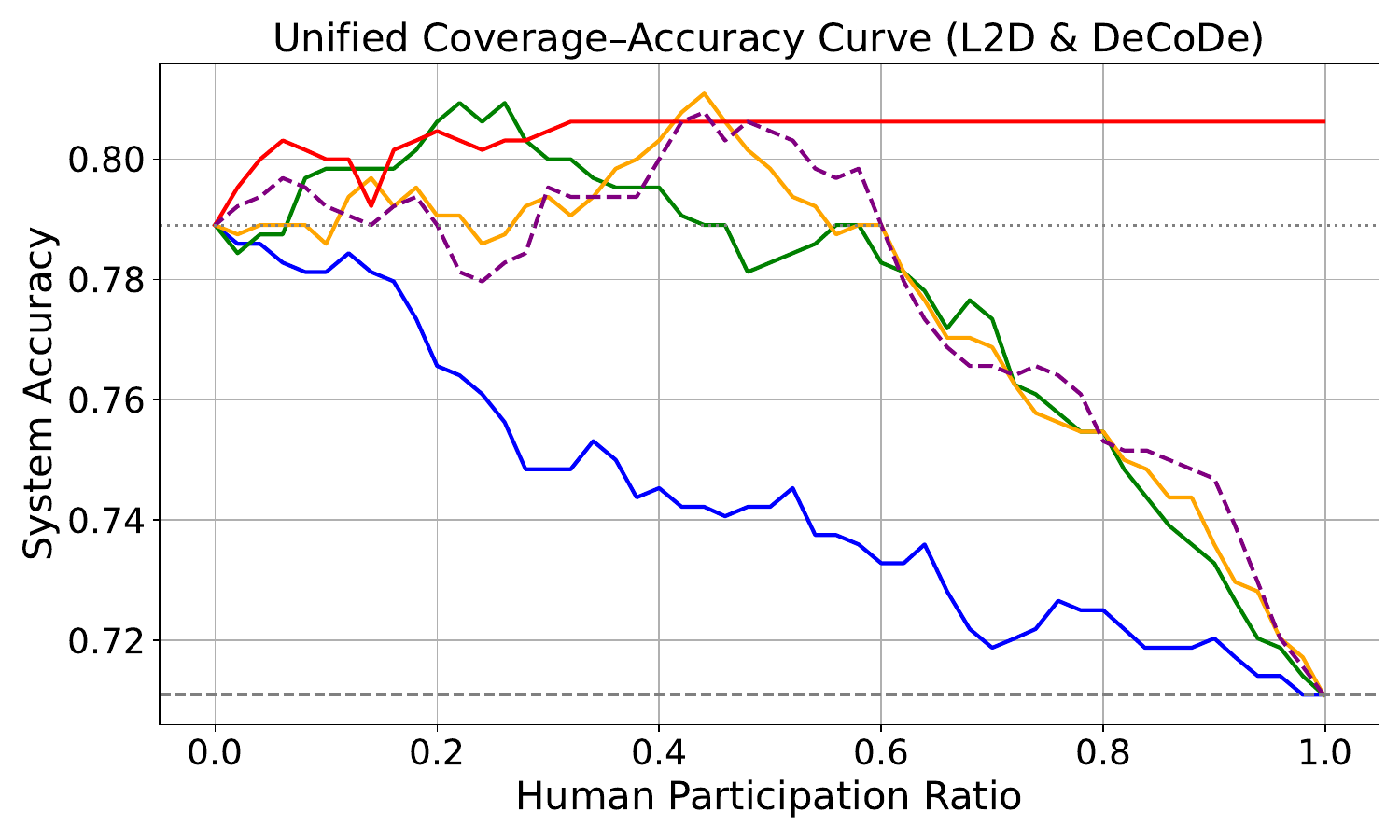}
  }
  \hspace{0.5em}
  \subfigure[Derm7pt, Noise 0.5 \label{fig:derm_noise_05}]{
    \includegraphics[width=0.3\linewidth]{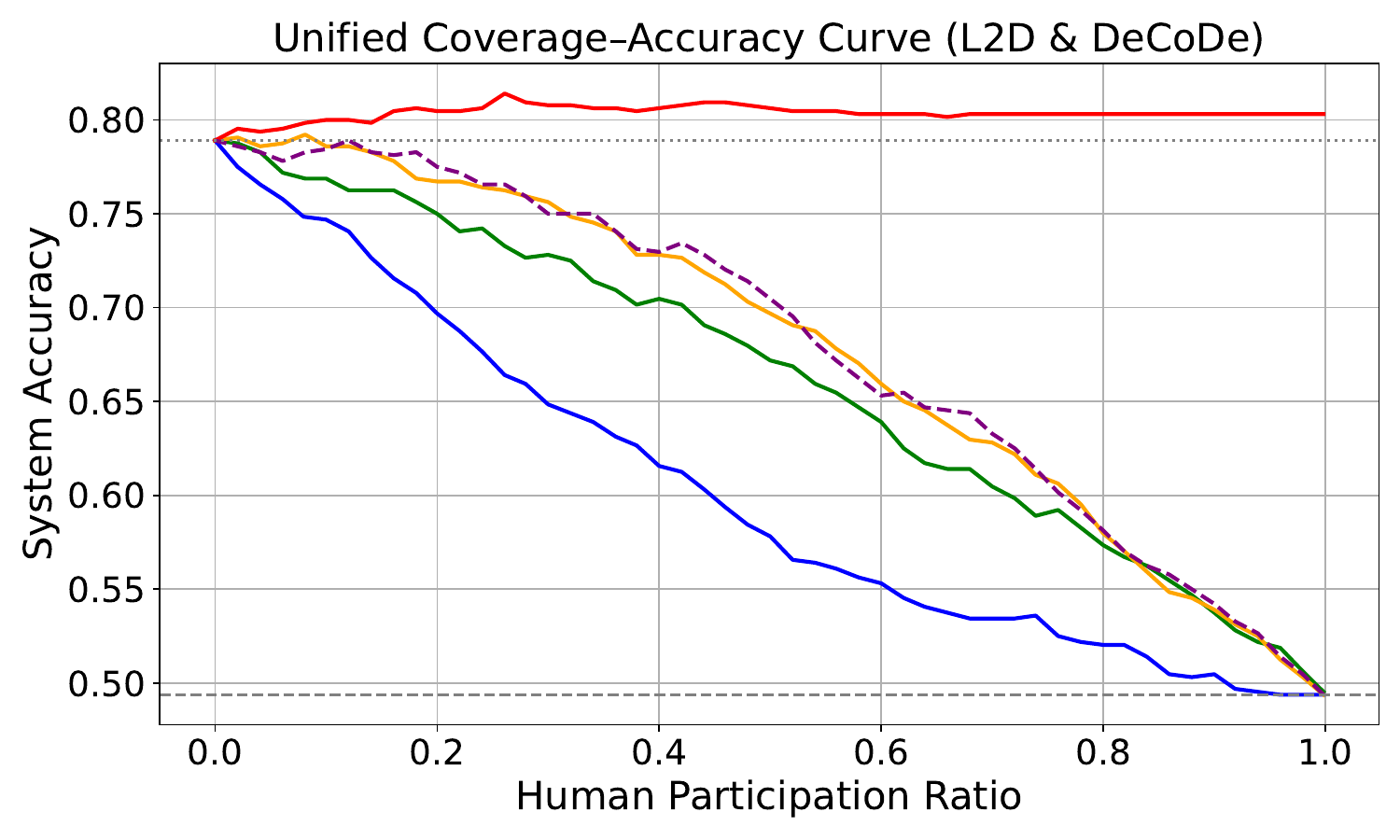}
  }
  \vspace{-0.5em}
  \subfigure[CelebA, Noise 0.1 \label{fig:celeba_noise_01}]{
    \includegraphics[width=0.3\linewidth]{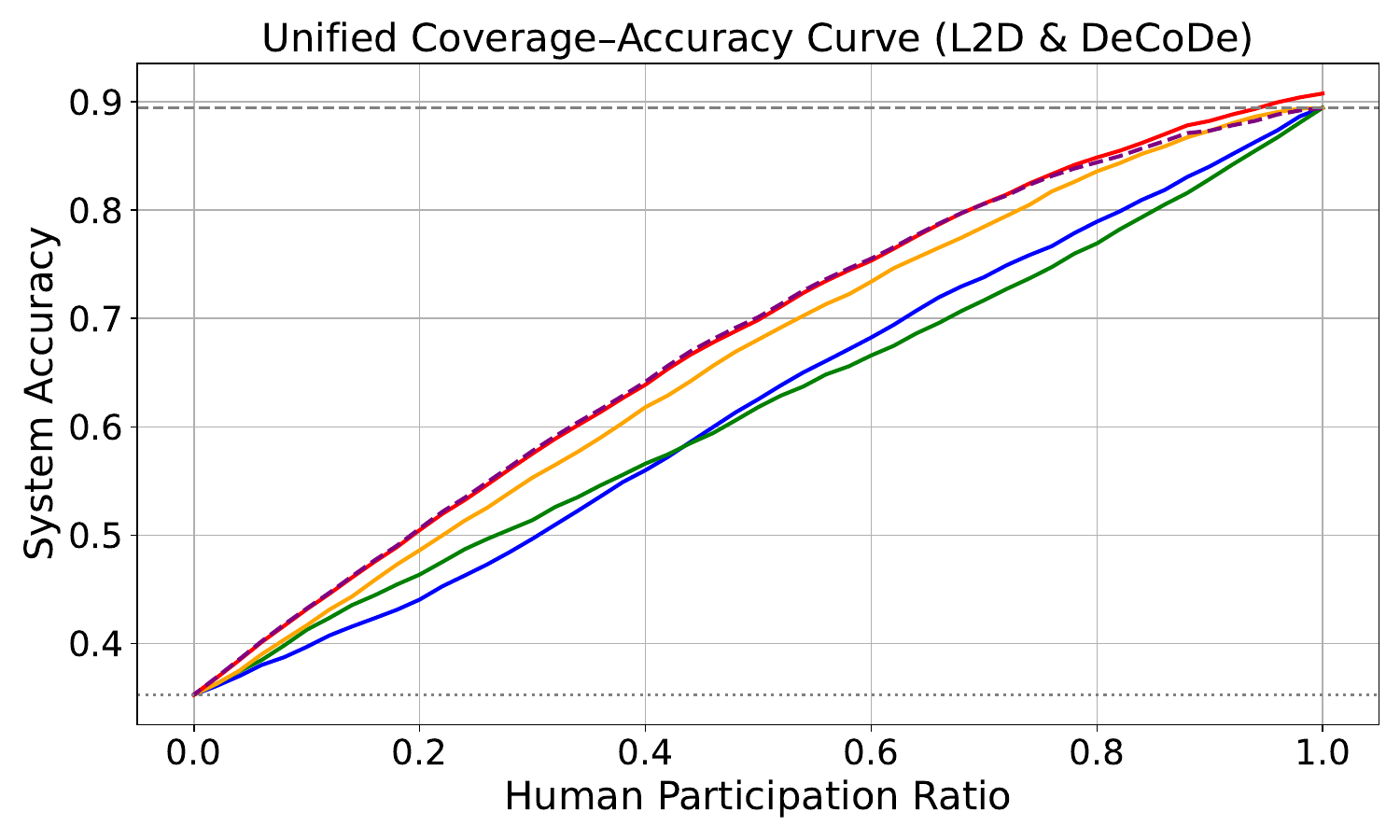}
  }
  \hspace{0.5em}
  \subfigure[CelebA, Noise 0.3 \label{fig:celeba_noise_03}]{
    \includegraphics[width=0.3\linewidth]{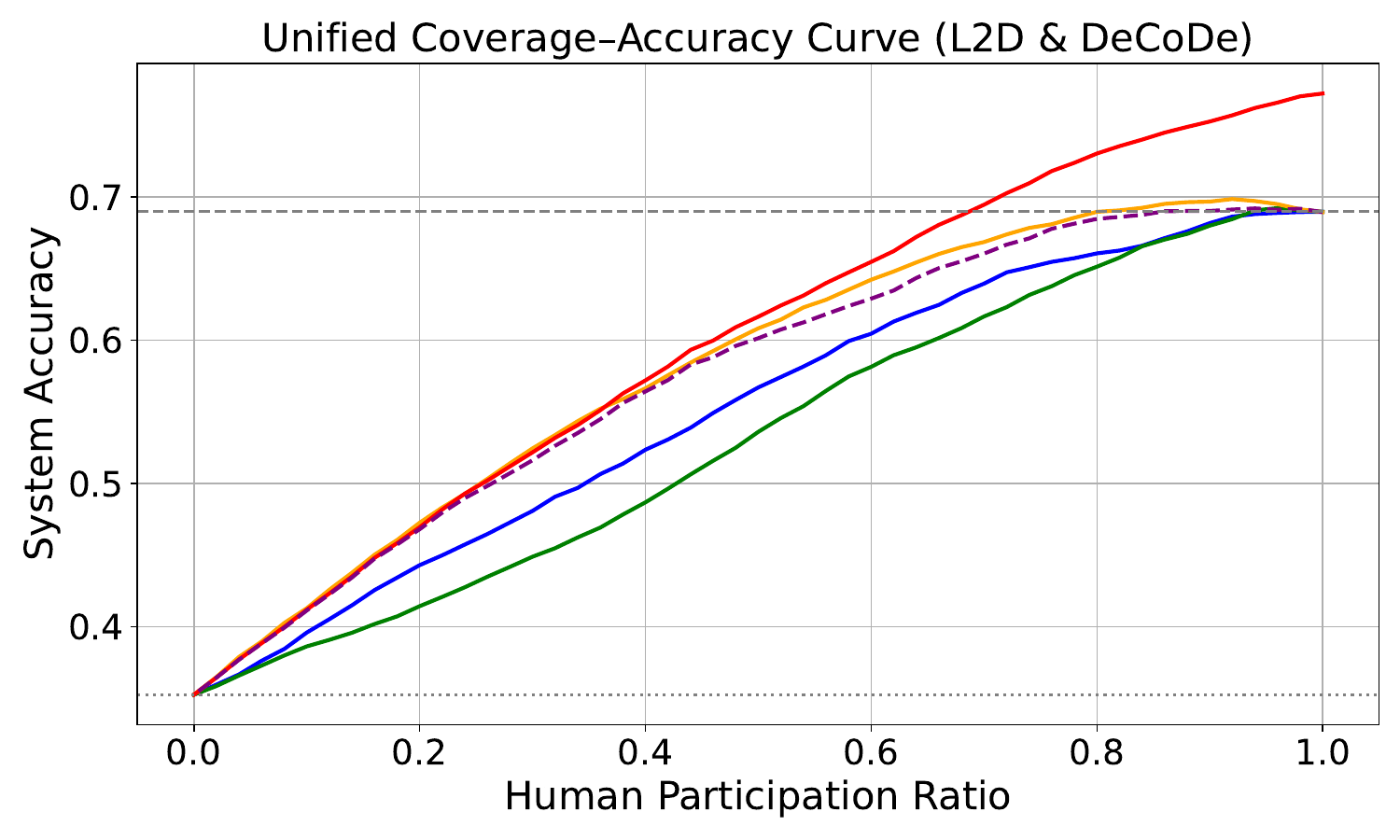}
  }
  \hspace{0.5em}
  \subfigure[CelebA, Noise 0.5 \label{fig:celeba_noise_05}]{
    \includegraphics[width=0.3\linewidth]{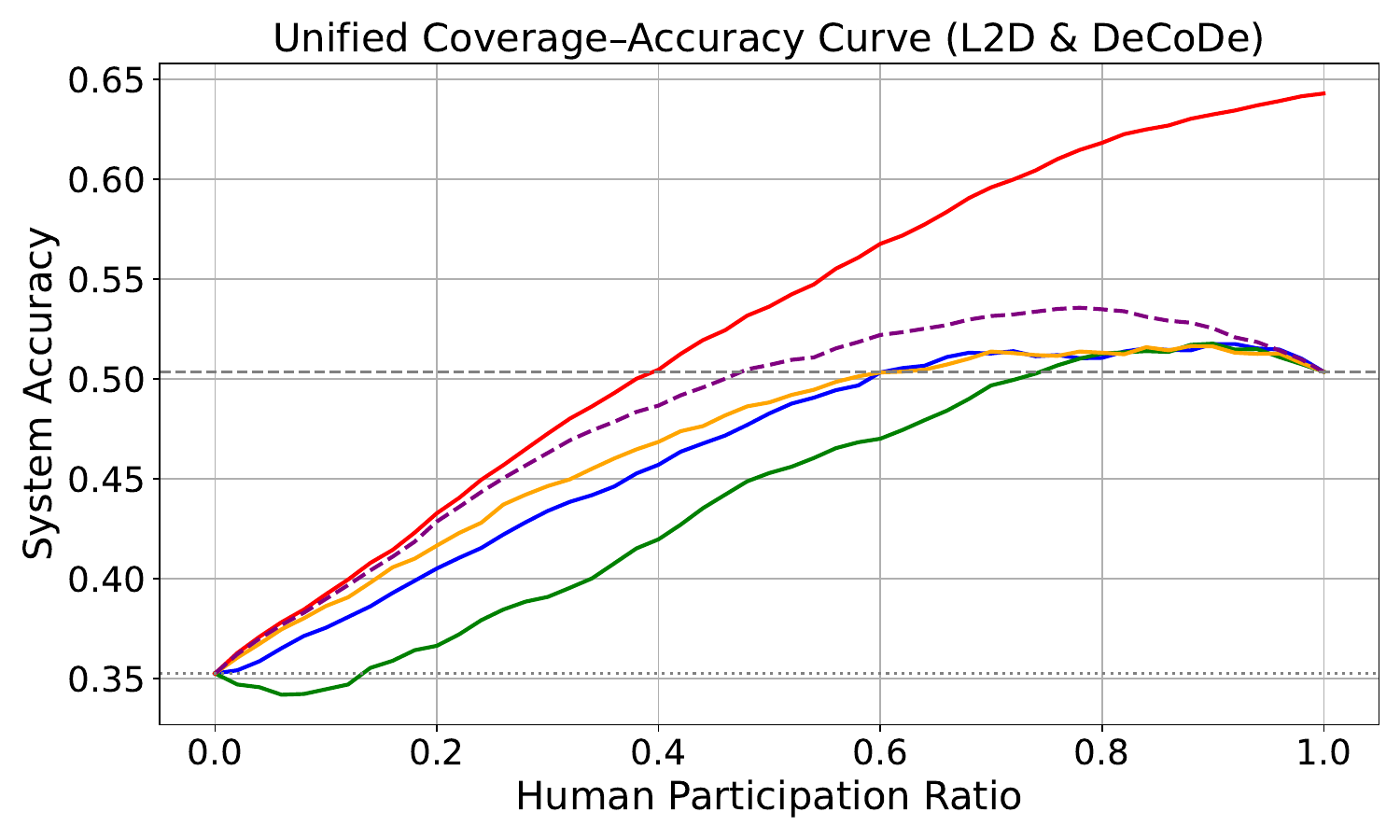}
  }
  \caption{System accuracy as a function of human participation ratio for \textbf{DeCoDe} and baseline methods under different noise levels. Each row corresponds to a dataset (CUB, Derm7pt, CelebA), and each column represents a simulated human noise rate (0.1, 0.3, 0.5). The x-axis denotes the proportion of test instances receiving human input, from 0 (AI-only) to 1 (Human-only). Dashed and dash-dotted lines indicate the standalone performance of human and AI agents, respectively.
}
  \label{fig:decode}
\end{figure}

DeCoDe is designed to operate effectively across various human reliability conditions. We examine concept and label prediction accuracy using a standard DCBM across the three datasets to illustrate differences in concept-label alignment. On CUB, concept accuracy reaches 96.86\% and label accuracy is 80.9\%; on Derm7pt, the results are 80.91\% and 78.9\%, respectively. In contrast, CelebA yields a high concept accuracy of 90.68\% but a much lower label accuracy of just 35.3\%. These discrepancies highlight that even highly accurate concept predictions do not always translate into reliable label inference, particularly when the alignment between concepts and class labels is weak. In such cases, human input remains essential to fill in missing contextual or semantic gaps. DeCoDe addresses this challenge by supporting flexible defer-and-complement strategies, enabling effective collaboration regardless of the underlying concept-label alignment. 

Figure~\ref{fig:decode} illustrates the accuracy–cost trade-offs of \textbf{DeCoDe} and several baselines on CUB, Derm7pt, and CelebA under varying simulated human label noise levels. Overall, DeCoDe consistently achieves superior performance and robustness across all datasets and noise levels, maintaining stable gains as human involvement increases and outperforming existing L2D methods in both low- and high-noise scenarios. In medium- and high-noise settings (Figure~\ref{fig:cub_noise_03}, \ref{fig:cub_noise_05}, \ref{fig:derm_noise_03}, \ref{fig:derm_noise_05}), DeCoDe shows increasing performance gains with higher human participation and effectively mitigates the impact of noisy supervision. Under low-noise conditions (e.g., Figure~\ref{fig:derm_noise_01}), it remains competitive with state-of-the-art methods. In contrast, variants like DeCoDe\_defer\_human—limited to binary AI-or-Human decisions—perform reasonably well at low cost but degrade significantly when human label quality deteriorates, indicating the limitations of rigid deferral under noisy expert input. In Figure~\ref{fig:cub_noise_01}, DeCoDe slightly lags behind L2D-REAL in early collaboration phases, likely because it emphasizes low-cost deferral signals through its surrogate loss. However, DeCoDe maintains greater stability as human participation increases, leading to stronger performance in the mid-to-high collaboration regime.

Interestingly, on CelebA, although concept prediction accuracy remains high, the corresponding label accuracy is substantially lower, revealing a gap between semantic features and final classification targets. Rather than weakening interpretability, this limitation highlights DeCoDe's ability to recognize when concept-level information is insufficient and explicitly defers such decisions to human experts. This boundary-aware deferral mechanism enhances performance while preserving transparency, constituting a core part of DeCoDe's interpretability.

Across all noise levels on CelebA (Figure~\ref{fig:celeba_noise_01}–\ref{fig:celeba_noise_05}), DeCoDe is the only method that continues to benefit from increased human involvement, while all L2D baselines plateau or decline. These results demonstrate that static deferral strategies are inadequate for real-world human-AI collaboration and that DeCoDe's flexible defer-and-complement mechanism enables more adaptive, robust, and interpretable decision-making. More experimental results can be found in Appendix D.

\subsection{Concept Intervenability via Test-Time Intervention}

\label{Concept Intervenability via Test-Time Intervention}

\begin{figure}[t]
  \centering
  \includegraphics[width=1\linewidth]{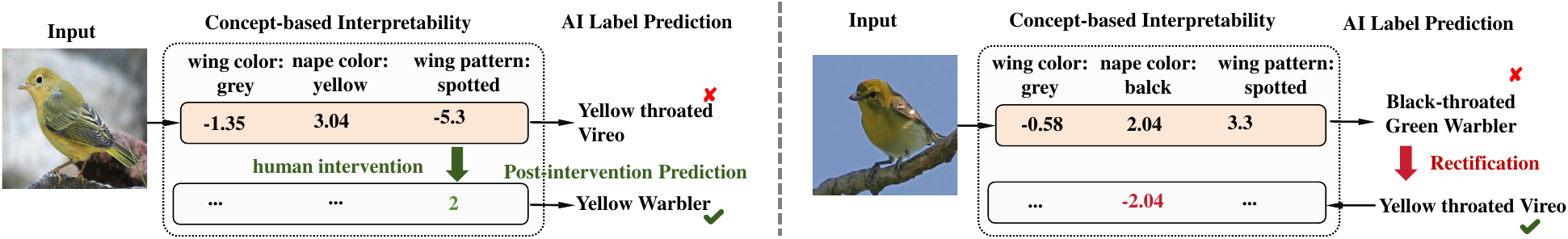}
    \caption{
        \textbf{Effect of concept-level intervention on downstream predictions.}
        The figure illustrates how manually modifying intermediate concepts can alter the model's final prediction. The left shows the input image, the middle shows predicted and intervened concept values, and the right shows label predictions before and after intervention. 
  }
  \label{fig:intervention}
\end{figure}

We evaluate DeCoDe's capacity for concept-level intervention, enabling users to easily influence predictions via semantic edits. DeCoDe supports both \textbf{forward intervention}—modifying incorrect concepts while holding the label fixed—and \textbf{backward rectification}—adjusting concept predictions given a corrected label.

In \textbf{forward intervention}, concept logits are replaced with high-confidence values (e.g., 95th/5th percentile) to simulate expert input. For mutually exclusive concept groups (e.g., CUB wing colors), edits are applied jointly, with visibility constraints enforced~\citep{koh2020concept, zhang2024decoupling}. As shown in Figure~\ref{fig:intervention} (left), minor concept edits can significantly revise the model output.

In \textbf{backward rectification}~\citep{zhang2024decoupling}, DeCoDe revises inconsistent concepts when the actual label is known. Despite low concept error rates ($<4\%$), such corrections improve semantic alignment, as demonstrated in Figure~\ref{fig:intervention} (right).

These mechanisms highlight DeCoDe's interactive, interpretable, and adaptive behavior, which are critical for effective human-AI collaboration. By supporting both forward intervention and backward rectification, DeCoDe enables real-time corrections, ensuring the system adapts to varying levels of human input and expertise.

\begin{figure}[t]
  \centering
  \subfigure[Coverage–Accuracy curves: DeCoDe (Concept) vs. DeCoDe (Image)\label{fig:policy_compare}]{
    \includegraphics[width=0.45\linewidth]{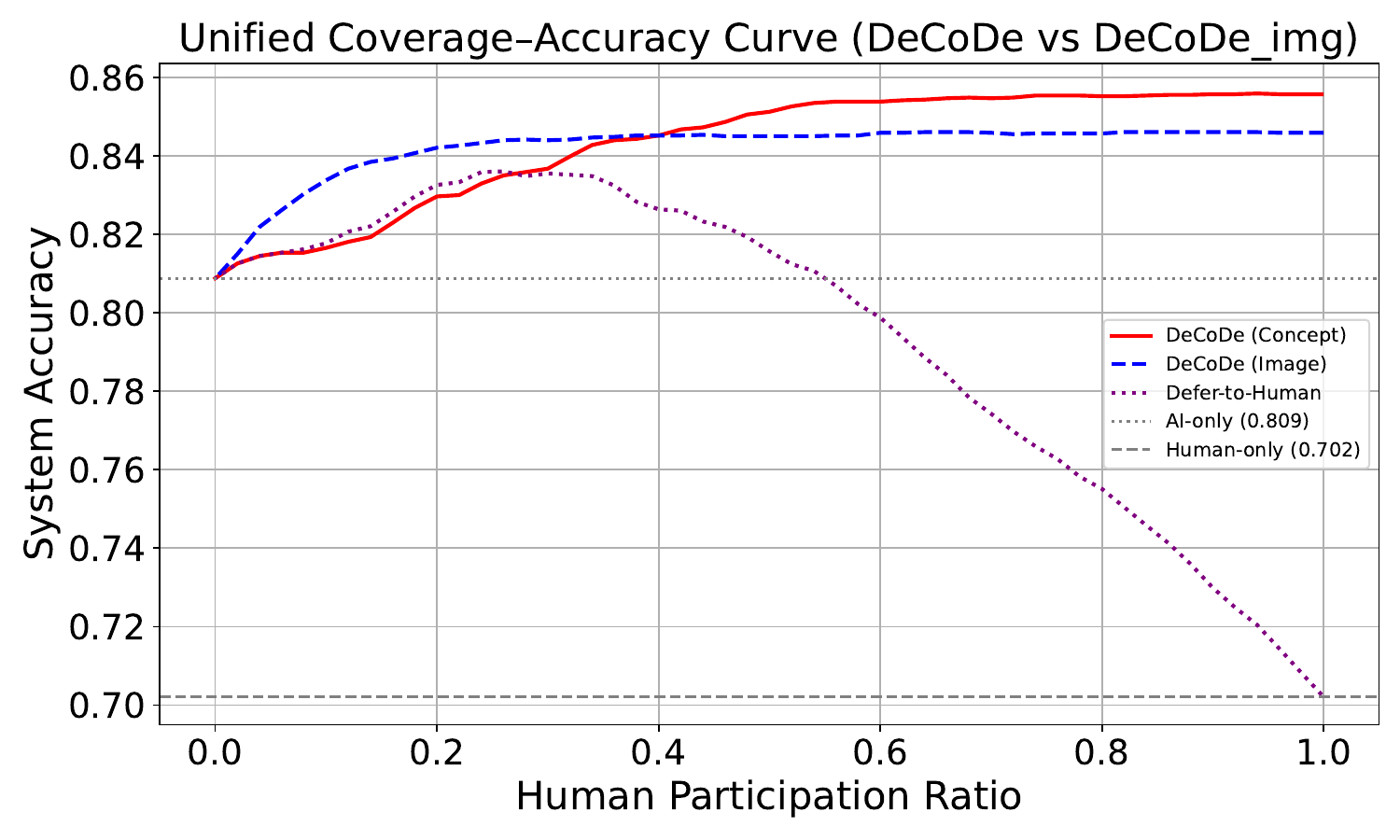}
  }\hfill
  \subfigure[Concept activation across strategies (soft aggregation)\label{fig:concept_strategy_heatmap}]{
    \includegraphics[width=0.45\linewidth]{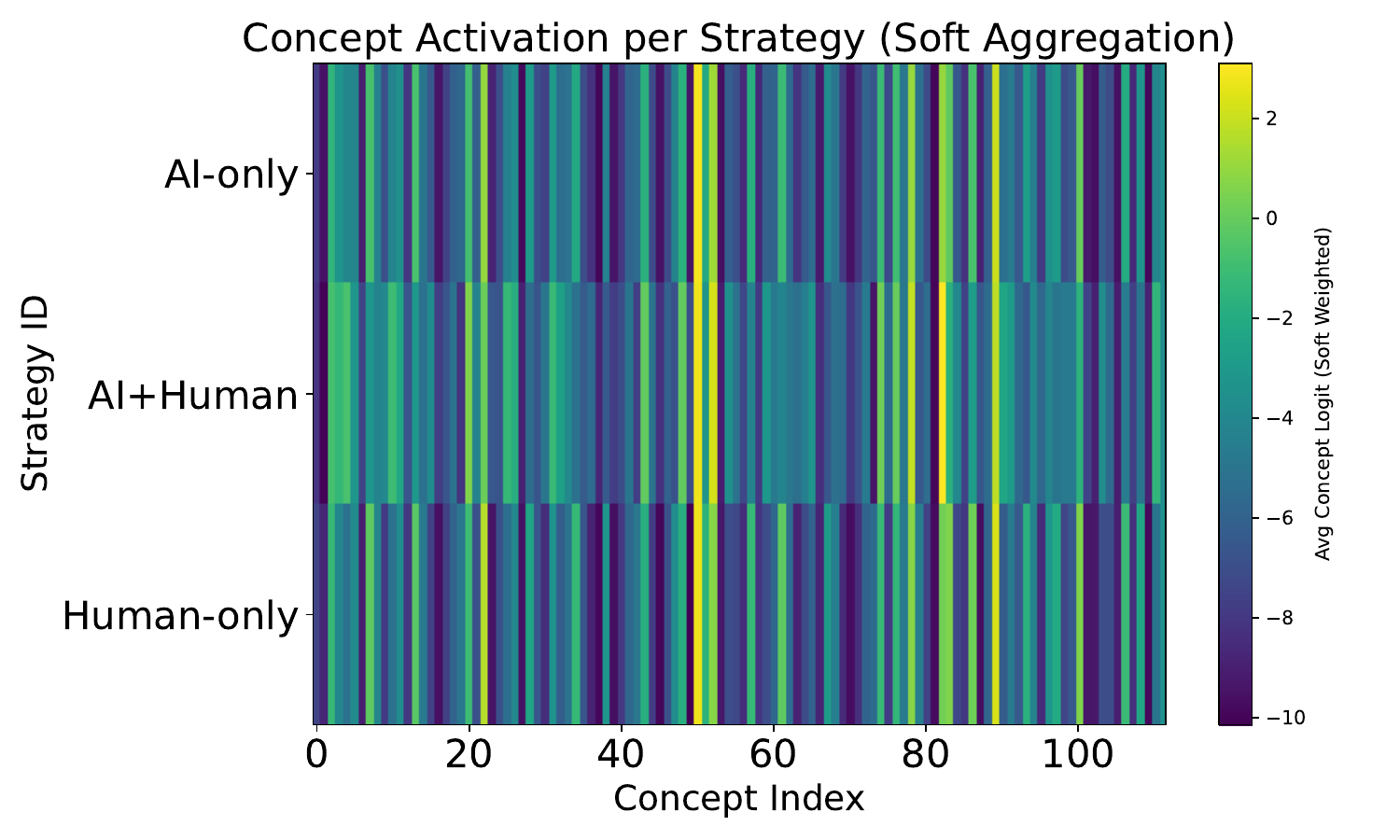}
  }
  \caption{
Analysis of concept-based strategy selection in \textsc{DeCoDe}. 
(a) Coverage–accuracy curves on the CUB dataset under simulated 70\% human accuracy. Concept-based gating (red) achieves better stability and overall accuracy as collaboration increases, while image-based gating (blue) performs slightly better at low coverage. 
(b) Heatmap of soft-aggregated concept activations across strategies, revealing distinct semantic preferences for AI-only, AI+Human, and Human-only paths. These results illustrate DeCoDe’s semantic adaptability in selecting collaboration modes.
}
  \label{fig:concept_strategy_analysis}
\end{figure}

\section{Discussion}
\label{Discussion}

Figure \ref{fig:policy_compare} compares two gating mechanisms in DeCoDe (image-based and concept-based) on the CUB dataset under a simulated setting with 70\% human accuracy. While DeCoDe (Image) slightly outperforms at low human involvement, DeCoDe (Concept) exhibits higher stability and overall accuracy as collaboration increases, particularly in high-coverage regions. These results underscore the benefits of concept-driven decision-making, which allows the model to adaptively defer or collaborate under uncertainty, rather than defaulting to autonomous predictions. This flexibility leads to improved robustness and interpretability in human-AI collaboration. Further experimental details on other datasets can be found in Appendix E.

Additionally, Figure \ref{fig:concept_strategy_heatmap} visualizes the aggregated concept activations for each strategy, weighted by strategy probabilities. We observe a clear semantic division of labor: the AI-only path emphasizes high-confidence, discriminative concepts, whereas the Human-only and AI+Human strategies rely more on perceptually salient, human-aligned attributes. This suggests DeCoDe can dynamically adapt strategy selection within a structured concept space based on semantic cues.

DeCoDe introduces an explicit concept bottleneck interface, enabling interpretable and traceable decision-making. Users can perform forward interventions by directly modifying concept predictions to influence outputs, while the model can also perform backward rectification by adjusting internal concept states based on label feedback. Together, these mechanisms form a bidirectional semantic loop. The GatingNet further integrates strategy selection into this semantic control pipeline, dynamically choosing between deferral, complement, or autonomous prediction based on concept-level input. This full-loop reasoning, from perception to strategy to intervention, enhances system transparency, controllability, and interaction, making DeCoDe particularly suitable for real-world applications such as medical diagnosis and educational support.

\section{Conclusion}
\label{Conclusion}

We propose \textbf{DeCoDe}, a concept-driven framework for human-AI collaboration that overcomes key limitations of existing L2D methods. Unlike traditional L2D approaches that make binary decisions between AI and humans, DeCoDe uses human-interpretable concept representations to support adaptive decisions in three modes: autonomous AI prediction, deferral to humans, and human-AI collaboration. A gating network, trained with a novel surrogate loss, selects the appropriate mode while balancing accuracy and human effort. By decoupling concepts from classification, DeCoDe enhances transparency and interpretability. Experiments on real-world datasets show that DeCoDe outperforms AI-only, human-only, and traditional deferral baselines, maintaining robustness and interpretability even under noisy human annotations. 

However, DeCoDe currently relies on manually annotated concepts, limiting its applicability in domains without structured supervision or labeled data. Future work will focus on automatic concept discovery and leveraging pre-trained vision-language models for weakly supervised learning. Additionally, extending DeCoDe to interactive settings, such as active learning and real-time feedback, will further enhance its adaptability in dynamic environments.

\bibliographystyle{plainnat} 
\bibliography{References}

\end{document}